\documentclass{llncs}

\usepackage{amsmath}
\usepackage{amssymb}
\usepackage{amsfonts}

\usepackage{bbm}
\usepackage{enumerate}

\usepackage{color}
\usepackage{cite}
\usepackage{graphicx}
\usepackage{pstricks}
\usepackage{amssymb}
\usepackage{amsfonts}
\usepackage{amsmath}
\usepackage[amsmath,thmmarks]{ntheorem}

\definecolor{mygreen}{rgb}{.0,.8,.0}

\begin{document}

\title{Exploring the similarity of medical imaging classification problems}


\author{Veronika Cheplygina\inst{1,2}, Pim Moeskops\inst{1}, Mitko Veta\inst{1,3}, Behdad Dashtbozorg\inst{1}, Josien P.W. Pluim\inst{1,3}}

\institute{Medical Image Analysis, Department of Biomedical Engineering, Eindhoven University of Technology, The Netherlands
\and Biomedical Imaging Group Rotterdam, Departments of Radiology and Medical Informatics, Erasmus Medical Center, The Netherlands
\and Image Sciences Institute, University Medical Center Utrecht, The Netherlands
}


\maketitle

\begin{abstract}
Supervised learning is ubiquitous in medical image analysis. In this paper we consider the problem of meta-learning -- predicting which methods will perform well in an unseen classification problem, given previous experience with other classification problems. We investigate the first step of such an approach: how to quantify the similarity of different classification problems. We characterize datasets sampled from six classification problems by performance ranks of simple classifiers, and define the similarity by the inverse of Euclidean distance in this meta-feature space. We visualize the similarities in a 2D space, where meaningful clusters start to emerge, and show that the proposed representation can be used to classify datasets according to their origin with 89.3\% accuracy. These findings, together with the observations of recent trends in machine learning, suggest that meta-learning could be a valuable tool for the medical imaging community.
\end{abstract}

\section{Introduction}

Imagine that you are a researcher in medical image analysis, and you have experience with machine learning methods in applications A, B, and C. Now imagine that a colleague, who just started working on application D, asks your advice on what type of methods to use, since trying all of them is too time-consuming. You will probably ask questions like ``How much data do you have?'' to figure out what advice to give. In other words, your perception of the similarity of problem D with problems A, B and C, is going to influence what kind of ``rules of thumb'' you will tell your colleague to use.

In machine learning, this type of process is called \emph{meta-learning}, or ``learning to learn''. In this meta-learning problem, the samples are the different datasets A, B and C, and the labels are the best-performing methods on each dataset. Given this data, we want to know what the best-performing method for D will be. The first step is to characterize the datasets in a meta-feature space. The meta-features can be defined by properties of the datasets, such as sample size, or by performances of simple classifiers~\cite{vilalta2002perspective,DuiPekTax2004,cheplygina2015characterizing}. Once the meta-feature space is defined, dataset similarity can be defined to be inversely proportional to the Euclidean distances within this space, and D can be labeled, for example, by a nearest neighbor classifier.

Despite the potential usefulness of this approach, the popularity of meta-learning has decreased since its peak around 15 years ago. To the best of our knowledge, meta-learning is not widely known in the medical imaging community, although methods for predicting the quality of registration\cite{muenzing2014dirboost} or segmentation\cite{gurari2016pull} can be considered to meta-learn within a single application. In part, meta-learning seems less relevant today because of the superior computational resources. However, with the advent of deep learning and the number of choices to be made in terms of architecture and other parameters, we believe that meta-learning is worth revisiting in the context of its use in applications in medical image analysis.

In this paper we take the first steps towards a meta-learning approach for classification problems in medical image analysis. More specifically, we investigate the construction of a meta-feature space, where datasets known to be similar (i.e. sampled from the same classification problem), form clusters. We represent 120 datasets sampled from six different classification problems by performances of six simple classifiers and propose several methods to embed the datasets into a two-dimensional space. Furthermore, we evaluate whether a classifier is able to predict which classification problem a dataset is sampled from, based on only a few normalized classifier performances. Our results show that even in this simple meta-feature space, clusters are beginning to emerge and 89.3\% of the datasets can be classified correctly. We conclude with a discussion of the limitations of our approach, the steps needed for future research, and the potential value for the medical imaging community.

\section{Methods}\label{sec:methods}

In what follows, we make the distinction between a ``classification problem'' - a particular database associated with extracted features, and a ``dataset'' - a subsampled version of this original classification problem.

We assume we are given datasets $\{(D_i, M_i)\}_{1}^{n}$, where $D_i$ is a dataset from some supervised classification problem, and $M_i$ is a meta-label that reflects some knowledge about $D_i$. For example, $M_i$ could be the best-performing (but time-consuming) machine learning method for $D_i$. For this initial investigation, $M_i$ is defined as the original classification problem $D_i$ is sampled from.

We represent each dataset $D_i$ by performances of $k$ simple classifiers. Each of the $n$ $D_i$s is therefore represented by an $k$-dimensional vector $\mathbf{x}_i$ which together form a $n \times k$ meta-dataset $A_k$. Due to different inherent class overlap, the values in $A_k$ might not be meaningful to compare to each other. To address this problem, we propose to transform the values of each  $\mathbf{x}_i$ by:
\begin{itemize}
\item Normalizing the values to zero mean and unit variance, creating a meta-dataset $N_k$
\item Ranking the values between 1 and $k$, creating a meta-dataset $R_k$. In cases of ties, we use average ranks.
\end{itemize}

\begin{figure}
\centering
\includegraphics[width=0.9\textwidth]{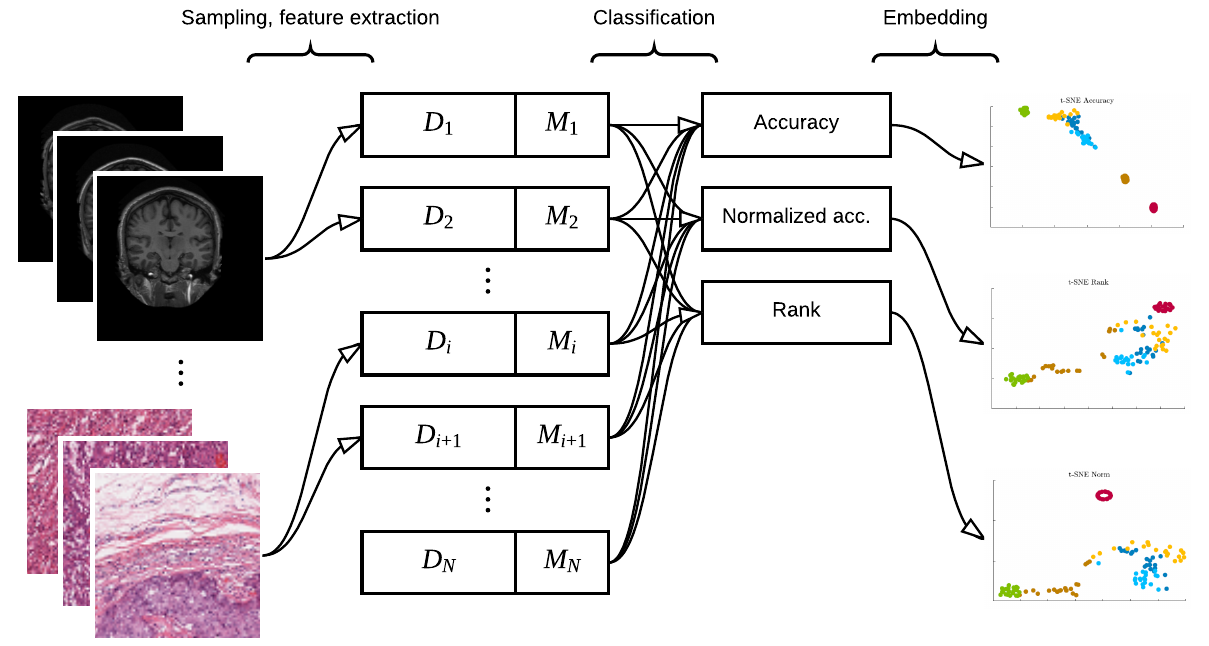}%
\caption{Overview of the method.}
\label{fig:overview}
\end{figure}

The final step is to embed the meta-datasets $A_k$, $N_k$ and $R_k$ in a 2D space for visualization, to obtain $A_2$, $N_2$ and $R_2$. We use two types of embedding: multi-dimensional scaling (MDS) \cite{cox2000multidimensional} and t-stochastic nearest neighbor embedding (t-SNE) \cite{maaten2008visualizing}. These embeddings can help to understand complementary properties of the data. MDS emphasizes large distances, and is therefore good at pointing out outliers, whereas t-SNE emphasizes small distances, potentially creating more meaningful visualizations. An overview of the approach is shown in Fig.~\ref{fig:overview}.

\section{Experiments}

\begin{table}
\centering
\caption{Classification problems described by type of image, number of images (subjects), number of classes and number and type of features: ``Classical'' = intensity and texture-based, ``CNN'' = defined by output of the fully-connected layer of a convolutional neural network.}
\begin{tabular}{l l l l l}
\hline
Dataset & Type & Images & Classes & Features \\
\hline

Tissue~\cite{Land12} & Brain MR & 20 & 7 & 768, CNN~\cite{Moes16} \\
Mitosis, MitosisNorm~\cite{Veta2015} & Histopathology & 12 & 2 & 200, CNN~\cite{veta2016} \\
Vessel~\cite{staal2004ridge} & Retinal & 20 & 2 & 29, Classical~\cite{zhang2016} \\
ArteryVein~\cite{dashtbozorg2014} & Retinal & 20 & 2 & 30, Classical~\cite{dashtbozorg2014} \\
Microaneurysm~\cite{Decenci2013} & Retinal & 381 & 2 & 30, Classical\\
\hline
\end{tabular}
\label{tab:datasets}
\end{table}

\noindent \textbf{Data and Setup.} We sample datasets from six classification problems, described in Table~\ref{tab:datasets}. The problems are segmentation or detection problems, the number of samples (pixels or voxels) is therefore much higher than the number of images. We sample each classification problem 20 times by selecting 70\% of the subjects for training, and 30\% of the subjects for testing, to generate $n=120$ datasets for the embedding. For each of dataset, we do the following:

\begin{itemize}
\item Subsample \{100, 300, 1K, 3K, 10K\} pixels/voxels from the training subjects
\item Train $k=6$ classifiers on each training set: nearest mean, linear discriminant, quadratic discriminant, logistic regression, 1-nearest neighbor and decision tree
\item Subsample 10K pixels/voxels from the test subjects
\item Evaluate accuracy on the test set, to obtain $5 \times 6$ accuracies
\item Transform the accuracies by ranking or normalizing
\item Average the ranks/normalized accuracies over the 5 training sizes
\end{itemize}

We use balanced sampling for both training and test sets, in order to keep performances across datasets comparable (and to remove the ``easy'' differences between the datasets). The classifiers are chosen mainly due to their speed and diversity (linear vs non-linear). The training sizes are varied to get a better estimation of each classifier's performance.

\smallskip
\noindent \textbf{Embedding.} We use the MDS algorithm with default parameters\footnote{http://prtools.org/}, and t-SNE\footnote{https://lvdmaaten.github.io/tsne/} with perplexity $=5$. Because of the stochastic nature of t-SNE, we run the algorithm 10 times, and select the embedding that returns the lowest error. We apply each embedding method to $A_k$, $N_k$ and $R_k$, creating embeddings $A_2^{tsne}$, $A_2^{mds}$ and so forth.

\smallskip
\noindent \textbf{Classification.} To quantify the utility of each embedding, we also perform a classification experiment. We train a 1-nearest neighbor classifier to distinguish between the different classification problems, based on the meta-representation. The classifiers are trained with a random subset of \{5, 10, 20, 40, 60\} meta-samples 5 times, and accuracy is evaluated on the remaining meta-samples.

\section{Results}\label{sec:results}

\begin{figure}
\centering
\includegraphics[width=0.95\textwidth]{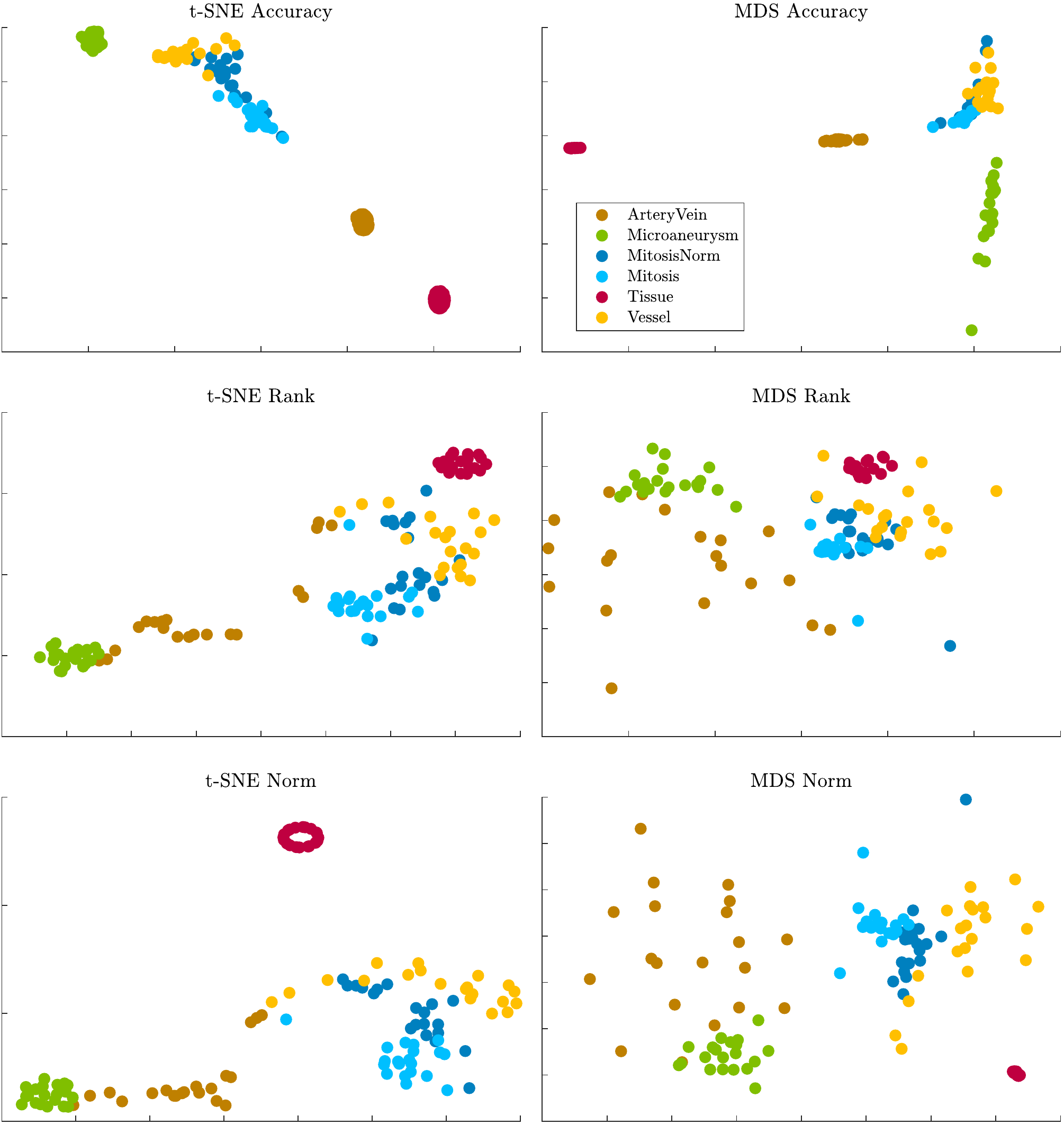}%
\caption{2D embeddings of the datasets with t-SNE (left) and MDS (right), based on accuracy (top), ranks (middle) and scaling (bottom).}
\label{fig:embedding}
\end{figure}


\noindent \textbf{Embedding.} The embeddings are shown in Fig.~\ref{fig:embedding}. The embeddings based on accuracy are the best at discovering the true structure. Although we sampled the datasets in a balanced way in an attempt to remove some differences in accuracy, it is clear that some problems have larger or smaller class overlap, and therefore consistently lower or higher performances. Both t-SNE and MDS are able to recover this structure.

Looking at $N_2$ and $R_2$, t-SNE appears to be slightly better at separating clusters of datasets from the same classification problem. This is particularly clear when looking at the ArteryVein datasets. Visually it is difficult to judge whether $N_2$ or $R_2$ provides a more meaningful embedding, which is why it is instructive to look at how a classifier would perform with each of the embeddings.

Looking at the different clusters, several patterns start to emerge. ArteryVein and Microaneurysm are quite similar to each other, likely to to the similarity of the images and the features used. Furthermore, Mitosis and MitosisNorm datasets are quite similar to each other, which is expected, because the same images but with different normalization are used. The Tissue dataset is often the most isolated from the others, being the only dataset based on 3D MR images.

Not all similarities can be explained by prior knowledge about the datasets. For example, we would expect the Vessel to be similar to the ArteryVein and Microaneurysm datasets, but it most embeddings it is actually more similar to Mitosis and MitosisNorm. This suggests that larger sample sizes and/or more meta-features are needed to better characterize these datasets.

\begin{figure}
\centering
\includegraphics[width=0.45\textwidth]{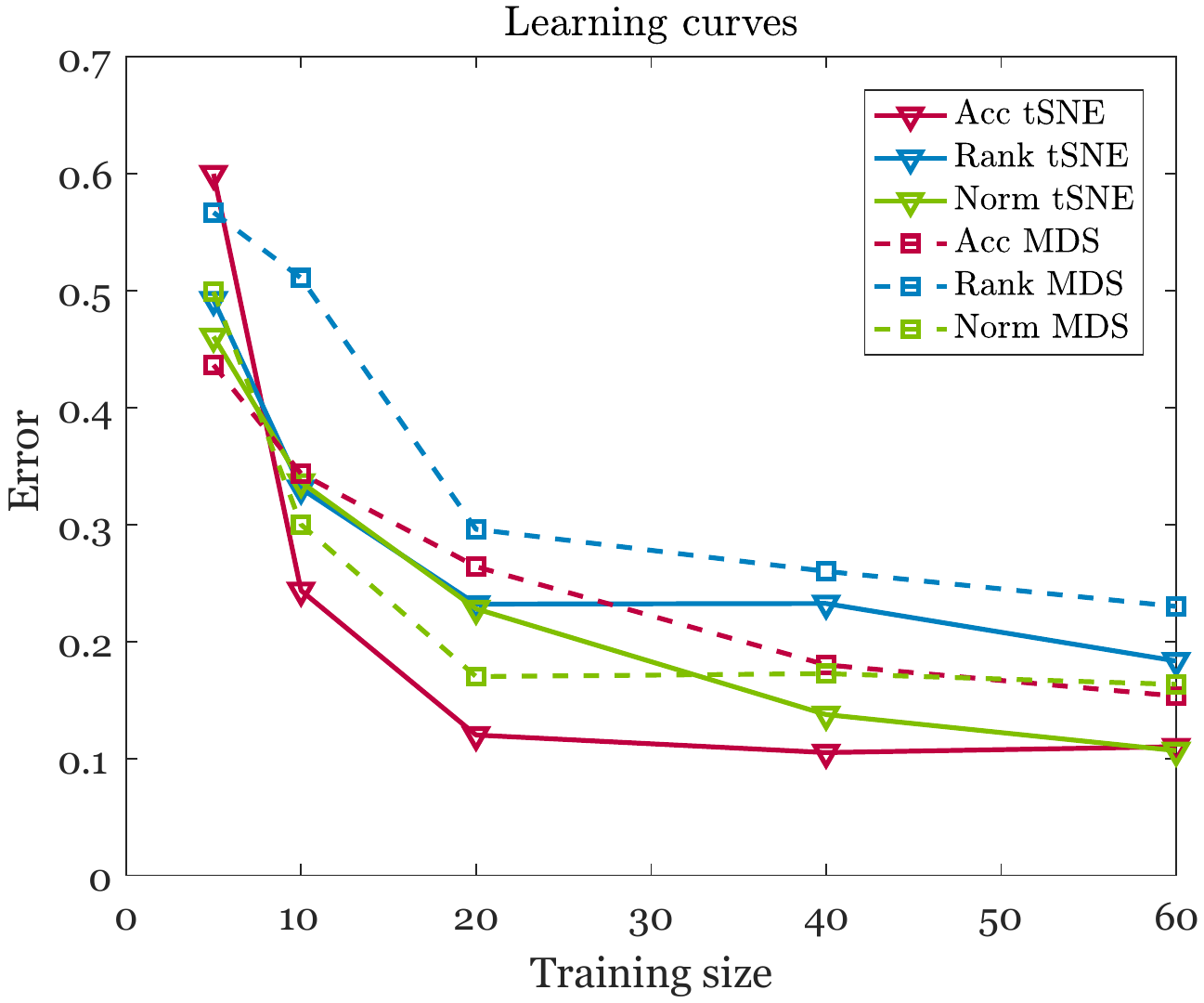}%
\includegraphics[width=0.45\textwidth]{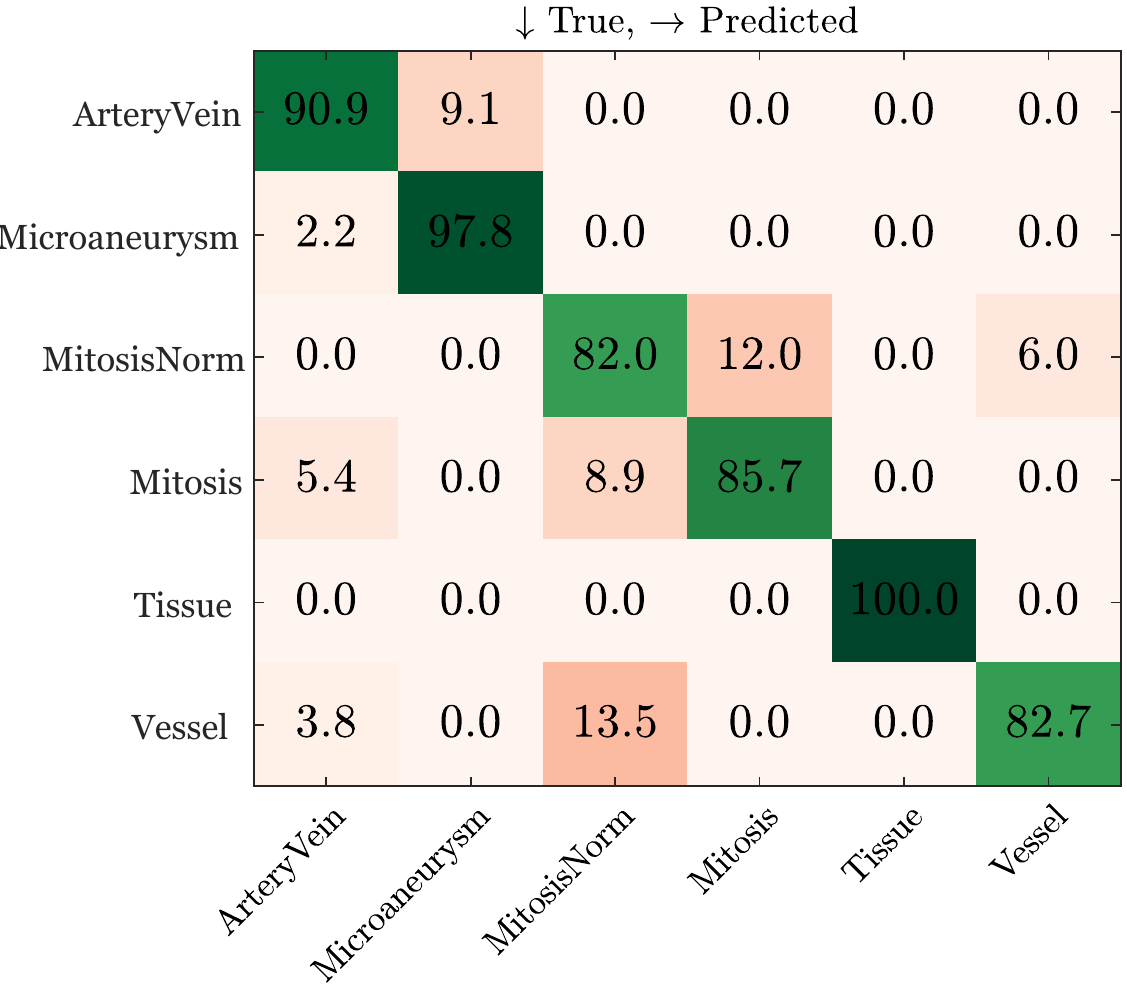}%
\caption{Left: Learning curves of 1-NN classifiers trained on different-size samples from six meta-datasets. Right: confusion matrix of classifier trained on 60 samples from $N_2^{tsne}$. Each cell shows what \% of the true class (row) is classified as (column).}
\label{fig:learncurve}
\end{figure}

\smallskip
\noindent \textbf{Classification.} The results of the 1-NN classifier, trained to distinguish between different classification problems, are shown in Fig.~\ref{fig:learncurve} (left). The performances confirm that the t-SNE embeddings are better than MDS.  As expected, $A_2$ is the best embedding. Between $N_2$ and $R_2$, which were difficult to assess visually, $N_2$ performs better. When trained on half (=60) of the meta-samples, 1-NN misclassifies 10.7\% of the remaining meta-samples, which is low for a 6-class classification problem.

To assess which samples are misclassified, we examine the confusion matrix of this classifier in Fig.~\ref{fig:learncurve} (right). Most confusion can be found between Mitosis and MitosisNorm, ArteryVein and Microaneyrism and between Vessel and the Mitosis datasets, as would be expected from the embeddings.

\section{Discussion and Conclusions}\label{sec:conclusions}

We presented an approach to quantify the similarity of medical imaging datasets, by representing each dataset by performances of simple classifiers. Even though we used small samples from each dataset and only six simple classifiers, this representation was reasonably successful (89.3\% accuracy) in predicting the origin of each dataset. This demonstrates the potential of using this representation in a meta-learning approach, with the goal of predicting which machine learning method will be effective for a previously unseen dataset.

A limitation of our experiments is using artificial meta-labels, based on each dataset's original classification problem. In a complete meta-learning approach, the labels would reflect the best-performing method on the dataset. However, we would not expect the best-performing method to change if a different subset of subjects is used (which is how the datasets sampled from the same classification problem differ from each other). Therefore, since we observe clusters with these artificial meta-labels, we also expect to observe clusters if more realistic meta-labels are used. Validating the approach with classifier-based meta-labels is the next step for future research in this direction.

Furthermore, we considered features as immutable properties of the classification problem. By considering the features as fixed, our approach would only be able to predict which classifier to apply to these already extracted features. However, due to recent advances in CNNs, where no explicit distinction is made between feature extraction and classification, we would want to start with the raw images. A challenge that needs to be addressed is how to represent the datasets at this point: for example, performances of classifiers on features extracted by CNNs pretrained on external data, or by some intrinsic, non-classifier-based characteristics.

Despite these limitations, we believe this study reaches a more important goal: that of increasing awareness about meta-learning, which is largely overlooked by the medical imaging community. One opportunity is to use meta-learning jointly with transfer learning or domain adaptation, which have similar goals of transferring knowledge from a source dataset to a target dataset. For example, pretraining a CNN on the source, and extracting features on the target, is a form of transfer. In this context, meta-learning could be used to study which source datasets should be used for the transfer: for example, a single most similar source, or a selection of several, diverse sources.

Another opportunity is to learn more general rules of thumb for ``what works when'' by running the same feature extraction and classification pipelines on different medical imaging datasets, such as challenge datasets. In machine learning, this type of comparison is already being facilitated by OpenML~\cite{vanschoren2014openml}, a large experiment database allows running different classification pipelines on already extracted features. We believe that a similar concept for medical imaging, that would also include preprocessing and feature extraction steps, would be a valuable resource for the community.



\bibliographystyle{splncs}
\bibliography{refs}

\begin{thebibliography}{10}

\bibitem{vilalta2002perspective}
Vilalta, R., Drissi, Y.:
\newblock A {Perspective} {View} and {Survey} of {Meta}-{Learning}.
\newblock Artificial Intelligence Review \textbf{18} (2002)  77--95

\bibitem{DuiPekTax2004}
Duin, R.P.W., Pekalska, E., Tax, D.M.J.:
\newblock The characterization of classification problems by classifier
  disagreements.
\newblock In: International Conference on Pattern Recognition. Volume~1. (2004)
   141--143

\bibitem{cheplygina2015characterizing}
Cheplygina, V., Tax, D.M.J.:
\newblock Characterizing multiple instance datasets.
\newblock In: Similarity-Based Pattern Recognition. (2015)  15--27

\bibitem{muenzing2014dirboost}
Muenzing, S.E.A., van Ginneken, B., Viergever, M.A., Pluim, J.P.W.:
\newblock {DIRBoost}--an algorithm for boosting deformable image registration:
  Application to lung {CT} intra-subject registration.
\newblock Medical Image Analysis \textbf{18}(3) (2014)  449--459

\bibitem{gurari2016pull}
Gurari, D., Jain, S.D., Betke, M., Grauman, K.:
\newblock Pull the {Plug}? {Predicting} {If} {Computers} or {Humans} {Should}
  {Segment} {Images}.
\newblock In: Computer Vision and Pattern Recognition. (2016)  382--391

\bibitem{cox2000multidimensional}
Cox, T.F., Cox, M.A.:
\newblock Multidimensional scaling.
\newblock CRC Press (2000)

\bibitem{maaten2008visualizing}
van~der Maaten, L., Hinton, G.:
\newblock Visualizing data using {t-SNE}.
\newblock Journal of Machine Learning Research \textbf{9} (2008)  2579--2605

\bibitem{Land12}
Landman, B.A.,  et~al.:
\newblock {MICCAI} 2012 Workshop on Multi-Atlas Labeling.
\newblock CreateSpace Independent Publishing Platform (2012)

\bibitem{Moes16}
Moeskops, P., Viergever, M.A., Mendrik, A.M., de~Vries, L.S., Benders, M.J.,
  I\v{s}gum, I.:
\newblock Automatic segmentation of {MR} brain images with a convolutional
  neural network.
\newblock IEEE Transactions on Medical Imaging \textbf{35}(5) (2016)
  1252--1261

\bibitem{Veta2015}
Veta, M., Van~Diest, P.J., Willems, S.M., Wang, H., Madabhushi, A., Cruz-Roa,
  A., Gonzalez, F., Larsen, A.B., Vestergaard, J.S., Dahl, A.B.,  et~al.:
\newblock Assessment of algorithms for mitosis detection in breast cancer
  histopathology images.
\newblock Medical Image Analysis \textbf{20}(1) (2015)  237--248

\bibitem{veta2016}
Veta, M., van Diest, P.J., Jiwa, M., Al-Janabi, S., Pluim, J.P.W.:
\newblock Mitosis counting in breast cancer: Object-level interobserver
  agreement and comparison to an automatic method.
\newblock PloS ONE \textbf{11}(8) (2016)  e0161286

\bibitem{staal2004ridge}
Staal, J., Abr{\`a}moff, M.D., Niemeijer, M., Viergever, M.A., van Ginneken,
  B.:
\newblock Ridge-based vessel segmentation in color images of the retina.
\newblock IEEE Transactions on Medical Imaging \textbf{23}(4) (2004)  501--509

\bibitem{zhang2016}
Zhang, J., Dashtbozorg, B., Bekkers, E., Pluim, J.P.W., Duits, R., ter
  Haar~Romeny, B.M.:
\newblock Robust retinal vessel segmentation via locally adaptive derivative
  frames in orientation scores.
\newblock IEEE Transactions on Medical Imaging \textbf{35}(12) (2016)
  2631--2644

\bibitem{dashtbozorg2014}
Dashtbozorg, B., Mendon\c{c}a, A.M., Campilho, A.:
\newblock An automatic graph-based approach for artery/vein classification in
  retinal images.
\newblock IEEE Transactions on Image Processing \textbf{23}(3) (2014)
  1073--1083

\bibitem{Decenci2013}
Decenci{\`e}re, E.,  et~al.:
\newblock {TeleOphta}: Machine learning and image processing methods for
  teleophthalmology.
\newblock IRBM \textbf{34}(2) (2013)  196--203

\bibitem{vanschoren2014openml}
Vanschoren, J., Van~Rijn, J.N., Bischl, B., Torgo, L.:
\newblock Open{ML}: networked science in machine learning.
\newblock ACM SIGKDD Explorations Newsletter \textbf{15}(2) (2014)  49--60

\end{thebibliography}

\end{document}